\documentclass[twocolumn]{article}

\usepackage{PRIMEarxiv}
\usepackage[utf8]{inputenc}
\usepackage[T1]{fontenc}
\usepackage{times}
\usepackage{xcolor}
\definecolor{myblue}{RGB}{65,105,225}
\usepackage[
  breaklinks,
  colorlinks=true,
  linkcolor=myblue,
  citecolor=myblue,
  urlcolor=myblue,
  unicode
]{hyperref}
\usepackage{url}
\usepackage[numbers]{natbib}
\usepackage{booktabs}
\usepackage{amsfonts}
\usepackage{nicefrac}
\usepackage{microtype}
\usepackage{lipsum}
\usepackage{fancyhdr}
\usepackage{graphicx}
\graphicspath{{media/}}
\usepackage{multicol}  
\usepackage{multirow} 
\usepackage{amsmath} 
\newcommand{\blackhyperlink}[2]{\href{#1}{\textcolor{black}{#2}}}
\title{CDAN: Convolutional Dense Attention-guided Network for Low-light Image Enhancement
}

\author{
    {Hossein Shakibania\textsuperscript{$\dagger$}\qquad
    Sina Raoufi\textsuperscript{$\dagger$}\qquad
    Hassan Khotanlou}\textsuperscript{$\ddagger$}\\[3.5pt]
    Bu-Ali Sina University \\[2.5pt]
    {\tt \{h.shakibania, s.raoufi\}@eng.basu.ac.ir, khotanlou@basu.ac.ir}
}

\begin{document}
\setlength{\parindent}{1em}
\twocolumn[
    \maketitle
    \begin{abstract}
      Low-light images, characterized by inadequate illumination, pose challenges of diminished clarity, muted colors, and reduced details. Low-light image enhancement, an essential task in computer vision, aims to rectify these issues by improving brightness, contrast, and overall perceptual quality, thereby facilitating accurate analysis and interpretation. This paper introduces the Convolutional Dense Attention-guided Network (CDAN), a novel solution for enhancing low-light images. CDAN integrates an autoencoder-based architecture with convolutional and dense blocks, complemented by an attention mechanism and skip connections. This architecture ensures efficient information propagation and feature learning. Furthermore, a dedicated post-processing phase refines color balance and contrast. Our approach demonstrates notable progress compared to state-of-the-art results in low-light image enhancement, showcasing its robustness across a wide range of challenging scenarios. Our model performs remarkably on benchmark datasets, effectively mitigating under-exposure and proficiently restoring textures and colors in diverse low-light scenarios. This achievement underscores CDAN's potential for diverse computer vision tasks, notably enabling robust object detection and recognition in challenging low-light conditions. The code is available at \href{https://github.com/SinaRaoufi/CDAN}{https://github.com/SinaRaoufi/CDAN}
    \end{abstract}
    \vspace{0.5in} 
]
\footnotetext{\textsuperscript{$\dagger$}Equal Contribution}
\footnotetext{\textsuperscript{$\ddagger$}Corresponding Author}
\section{Introduction}
With the rapid growth of technology and its integration into our daily lives, the field of computer vision has gained significant attention. Computer vision, the science that enables computers to extract, analyze, and interpret information from images and multi-dimensional data, plays a critical role in many applications, such as autonomous vehicles, medical imaging, surveillance, robotics, and agriculture. A core prerequisite for these systems to function effectively is the quality of the input images. High-quality images provide clear, detailed information that can enhance the accuracy of the resulting analysis and interpretation.

However, one of the considerable challenges faced in computer vision is dealing with low-light scenarios. Low-light images are those captured in inadequate light conditions, often resulting in poor visibility, increased noise, and loss of details. These images are problematic due to their low signal-to-noise ratio, reduced contrast, and color distortion. Low-light conditions are prevalent in many environments, including outdoor nighttime settings, indoor spaces with insufficient lighting, shadows, and backlighting, posing substantial challenges for image understanding and subsequent tasks \cite{Guo2023}.

Low-light image enhancement is a computer vision task to improve the quality and visibility of images captured in low-light conditions. It aims to rectify the issues caused by poor lighting, such as enhancing the brightness of the image, improving the contrast, and reducing noise without introducing additional visual artifacts or distortions. Improving image quality can significantly enhance the performance of subsequent computer vision tasks, such as object detection, recognition, and tracking.

Deep learning has shown considerable promise in addressing the challenges of low-light image enhancement. Deep learning algorithms, specifically Convolutional Neural Networks (CNNs), excel in handling 2D images. These algorithms learn hierarchical feature representations from the input data, identifying complex patterns that can be used to enhance image quality significantly. In the context of low-light image enhancement, deep learning models can be trained to understand the characteristics of low-light images and apply appropriate transformations to improve their quality. They can effectively increase brightness, enhance contrast, and reduce noise, even in complex and highly variable low-light scenarios. Moreover, Generative Adversarial Networks (GANs) \cite{goodfellow2014generative}, another deep learning architecture, have been instrumental in generating high-quality images from low-light inputs, maintaining the integrity of the original details while introducing minimal distortions. The use of deep learning in low-light image enhancement has thus demonstrated a significant improvement over traditional enhancement techniques, providing more accurate and reliable results and paving the way for more robust computer vision applications in challenging light conditions.

This paper introduces the Convolutional Dense Attention-guided Network (CDAN) as a novel and robust approach to low-light image enhancement. Based on an architecture inspired by autoencoders, CDAN combines convolutional and dense blocks to capture intricate patterns, enabling comprehensive feature learning. The seamless integration of skip connections substantially strengthens the network's coherent data propagation. Remarkably, CDAN stands apart by embracing attention mechanisms, notably the Channel Attention Module (CAM) and the Spatial Attention Module (SAM), enhancing the model's adaptability in aggregating critical information. During training, we employ a composite loss function that combines L2 and VGG losses, enhancing both the quantitative and qualitative aspects of our model's performance. A dedicated post-processing phase enhances the already promising model outputs, focusing on improving color balance and contrast. CDAN's remarkable performance is showcased through its state-of-the-art results in both quantitative and qualitative assessments, underscoring its superiority on benchmark datasets.

\section{Related works}
In recent years, low-light image enhancement techniques have attracted significant research attention, resulting in a multitude of proposed methods in this area. Low-light image enhancement techniques can be categorized into traditional methods, focusing on gray-level transformation, histogram equalization, and Retinex-based techniques \cite{Guo2023}, and deep learning-based approaches, including supervised, unsupervised, and zero-shot learning methods \cite{Li2022}.

\subsection{Traditional methods}
\textbf{Gray-level transformation methods.} Gray-level transformation techniques, such as logarithmic transformation and gamma correction, are fundamental tools in image processing used to adjust brightness and contrast. These techniques have found extensive applications in medical imaging \cite{Chaudhury2015, Zhao2016}, underwater photography \cite{Voronin2018, Mishra2023}, and color correction \cite{VELUCHAMY2019}. Although effective for image enhancement, especially in low-light conditions, gray-level transformations are often employed in combination with complementary techniques, such as histogram equalization and Retinex-based methods, to improve the quality further and achieve more sophisticated and robust results \cite{Xia2019, VELUCHAMY2019}.

\textbf{Histogram equalization methods.} Histogram equalization (HE) algorithms have been extensively investigated and utilized for image enhancement, resulting in the development of numerous algorithms based on this technique. Typically, HE-based methods partition the histogram and image into sub-components, followed by performing histogram equalization operations on each sub-component individually \cite{Guo2023}. Kim et al. \cite{Kim1997} suggested a Brightness-preserving Bi-Histogram Equalization (BBHE), which separates the histogram of the input image into two sub-histograms based on the image's mean intensity. The two sub-histograms are then separately equalized. While maintaining the input image's mean brightness, BBHE can improve visual contrast. The equal-area Dualistic Sub-Image Histogram Equalization (DSIHE) approach, which Wang et al. \cite{Wang1999} claim is inspired by BBHE, divides the picture into two equal-area sub-images based on the probability density function and then equalizes each sub-image separately. Despite their effectiveness in low-light image enhancement, HE methods may have a weakness in not considering the global context, which can result in over-enhancement or loss of important details, especially in images with complex lighting conditions or wide dynamic ranges.

\textbf{Retinex-based methods.} Retinex-based methods decompose images into illumination and reflectance components to enhance low-light images while preserving essential details and improving visibility \cite{Guo2023}. The Single-Scale Retinex (SSR), proposed by Jobson et al. \cite{Jobson1997}, served as the basis, but it could only achieve either dynamic range reduction or color/lightness rendition. Rahman et al. \cite{Rahman1996} proposed the Multi-Scale Retinex (MSR) algorithm, combining enhancement results at different scales and considering local and global information. MSR estimates the illumination component by combining central surround functions of varying scales. Jobson et al. \cite{JobsonMSRCR} later introduced Multi-Scale Retinex with Color Restoration (MSRCR), enabling simultaneous dynamic range compression, color consistency, and lightness rendition. Li et al. \cite{Li2015} introduced a Retinex algorithm that utilizes a recursive bilateral filter to tackle the problem of slow processing speed encountered in the bilateral Retinex method. Guo et al. \cite{Guo2017} proposed LIME, a method to enhance low-light images by estimating pixel illuminations and refining them using an Augmented Lagrangian Multiplier (ALM) algorithm with a sped-up solver to reduce the computational load. Ren et al. \cite{Ren2020} presented the Low-rank Regularized Retinex Model (LR3M), which enhances low-light images by incorporating low-rank priors to suppress noise in the reflectance map. Jia et al. \cite{JIA2023} introduced a robust Retinex-based model with reflectance map re-weighting, aiming to enhance brightness levels and achieve concurrent re-balancing in low-light images under non-uniform lighting conditions.

\subsection{Deep learning-based Methods}
\textbf{Supervised learning methods.} The first deep learning-based low-light image enhancement method, LLNet, proposed by Lore et al. \cite{LORE2017}, utilized a supervised learning (SL) based approach with an autoencoder to perform contrast enhancement and denoising tasks for low-light images jointly. Li et al. \cite{LI2018} introduced LightenNet, a Retinex-based CNN structure that predicts the illumination map for weakly illuminated images. The advantage of LightenNet lies in its ease of training and ability to establish mapping relations between the input images and their illumination maps. Lv et al. \cite{Lv2018MBLLEN} presented MBLLEN, a CNN-based method incorporating feature extraction, enhancement, and fusion modules to suppress noise and artifacts in low-light areas effectively. Wei et al. \cite{Wei2018Retinex} proposed Retinex-Net, which follows a three-step process involving decomposition, adjustment, and reconstruction for image enhancement. The Decom-Net separates the input into reflectance and illumination, while Enhance-Net brightens the illumination and reduces reflectance noise. Finally, the adjusted components are reconstructed for enhanced output. Wang et al. \cite{Wang2018} proposed GLADNet, an encoder-decoder network that estimates global illumination for low-light inputs and adjusts illumination accordingly while preserving details through a convolutional network. Ren et al. \cite{Ren2019} proposed a network with two streams, one for learning global content and the other for capturing salient structures of low-light images. To overcome the detail loss caused by the encoder, they suggested a spatially variant recurrent neural network (RNN) in the edge stream, guided by an additional auto-encoder. Zhang et al. \cite{Zhang2019} introduced KinD, a three-module network based on Retinex theory. The decomposition sub-module separates images into illumination and reflectance components, while the illumination sub-module handles light adjustment, and the reflectance sub-module focuses on removing degradation. Later, the authors introduced KinD++ \cite{Zhang2021}, an enhanced version of KinD specifically designed to remove hidden artifacts from images effectively. Zamir et al. \cite{Zamir2020} introduced MIRNet, an architecture with a multi-scale residual block that incorporates parallel multi-resolution convolution streams, information exchange, spatial and channel attention mechanisms, and attention-based multi-scale feature aggregation to maintain spatially-precise high-resolution representations and capture strong contextual information. In their study, Lu et al. \cite{lu2020tbefn} introduced the Two-Branch Exposure-Fusion Network (TBEFN). This network was designed to generate two enhancements which are then combined into a final output using a self-adaptive attention unit. Lim et al. \cite{Lim2021} proposed a method called Deep Stacked Laplacian Restorer (DSLR) that leverages the Laplacian pyramid in the input image and feature spaces to recover global illumination and local details separately,  combining them in the image space for enhanced efficiency. Lim et al. \cite{LIM2023} introduced Low-light Advanced U-Net (LAU-Net), an enhanced version of U-Net \cite{Ronneberger2015}, which integrates the Parallel Attention Unit (PAU), Internal Resizing Module (IRM), and external convolutional layers to enhance feature extraction and suppress noise. Makwana et al. \cite{makwana2024livenet} introduced LIVENet, which operates in two stages: image enhancement, where the Latent Subspace Denoising Block (LSDB) reduces noise and replaces the luminance channel with a noise-free grayscale image, and refinement, which integrates texture details using a Spatial Feature Transform (SFT) layer.

\textbf{Unsupervised learning methods.}
Unsupervised learning (UL) techniques in low-light image enhancement utilize unpaired datasets, where the model learns to enhance weakly illuminated images without access to ground truth data. Jiang et al. \cite{jiang2021} proposed EnlightenGAN, which employs a global-local discriminator structure, self-regularization with preserving loss, and an attention mechanism, achieving comparative results with supervised methods. Hu et al. \cite{hu2021} proposed a two-stage unsupervised method. The technique involves pre-enhancement using a conventional Retinex-based method, followed by post-refinement using an adversarially trained refinement network to improve image quality. Wang et al. \cite{Wang2022MAGAN} proposed MAGAN, a mixed-attention guided generative adversarial network, incorporating a mixed-attention module layer to model pixel-feature relationships for simultaneous enhancement and noise removal in low-light images.

\textbf{Semi-supervised learning method.}
Yang et al. \cite{Yang2021} introduced a Deep Recursive Band Network (DRBN) guided by pairs of low and normal-light images, enabling quality improvement through perceptually-driven linear transformations based on an image quality assessment neural network.

\textbf{Zero-shot learning methods.}
Zero-shot learning is a deep learning technique where a model can recognize and classify objects it has never seen during training. Zhang et al. \cite{Zhang2019Zero-shot} introduced ExCNet, an image-specific CNN, to restore low-light images by estimating the S-curve without relying on prior examples or training. Guo et al. \cite{Guo2020} introduced Zero-reference Deep Curve Estimation (Zero-DCE), a method utilizing deep network-based image-specific curve estimation for low-light image enhancement, employing four non-reference loss functions for zero-reference training. Zhu et al. \cite{Zhu2020} introduced RRDNet, a three-branch CNN, for denoising and restoring underexposed images through explicit noise prediction. Using a zero-shot learning approach, RRDNet decomposes input images into illumination, reflectance, and noise components. Liu et al. \cite{Liu2021} introduced RUAS, a technique that captures underexposed image structure through unrolling optimization processes based on the Retinex theory. Using a reference-free learning strategy, RUAS discovers effective architectures within a constrained search space, resulting in a high-performing and efficient enhancement network. Wang et al. \cite{quadprior} proposed a zero-reference framework for low-light enhancement using an illumination-invariant prior based on light transfer theory and a generative diffusion model. They introduce a bypass decoder to manage detail distortion and provide a lightweight version for practical applications. However, challenges such as robustness to diverse lighting conditions, color distortion, and maintaining visual fidelity remain critical for most of these methods.

Overall, traditional methods for low-light image enhancement are not feasible due to their vulnerability to complex lighting scenarios. Most newer methods are supervised and primarily based on U-Net architecture \cite{Lv2018MBLLEN, Zhang2019, lu2020tbefn, Lim2021, LIM2023, jiang2021, Guo2020}, but they still face limitations such as color distortions, poor color refinement, and inadequate texture preservation. Consequently, there is room for improvement. It is also noteworthy that few methods incorporate attention modules \cite{Zhang2019, Zamir2020, lu2020tbefn, LIM2023}, which help reduce information loss during feature extraction. Recently, there has been a rise in zero-shot learning and unsupervised approaches, which are more cost-efficient and do not rely on paired datasets. However, as shown in our comparisons and experiments, supervised methods generally yield better and more satisfying results.

\section{Methodology}
This study introduces a novel supervised autoencoder-based methodology that integrates CNNs, dense blocks, and attention modules, inspired by a comprehensive review of the literature and the proven efficiency of U-Net-based methods. Recognizing areas for improvement, we adopted an autoencoder-based model with skip connections to mitigate information degradation during the encoding process \cite{7780459}. Our model innovates by enabling dynamic information flow between encoder and decoder branches via dense blocks, a feature not seen in other state-of-the-art methods. We also incorporated an efficient, lightweight attention module at the autoencoder's bottleneck and within the decoder, significantly enhancing the preservation of important features during training, a technique used by only a few methods. This combination, derived from extensive experimentation and analysis, addresses limitations in existing methods, offering superior performance in low-light image enhancement. We further augment our approach with a post-processing stage to refine image quality. Our extensive quantitative and qualitative evaluations demonstrate the model's proficiency, making it a reliable tool for tasks such as detection, recognition, and scene understanding in extremely dark environments.

\subsection{Model architecture}
\begin{figure*}
    \centering
    \includegraphics[width=\linewidth]{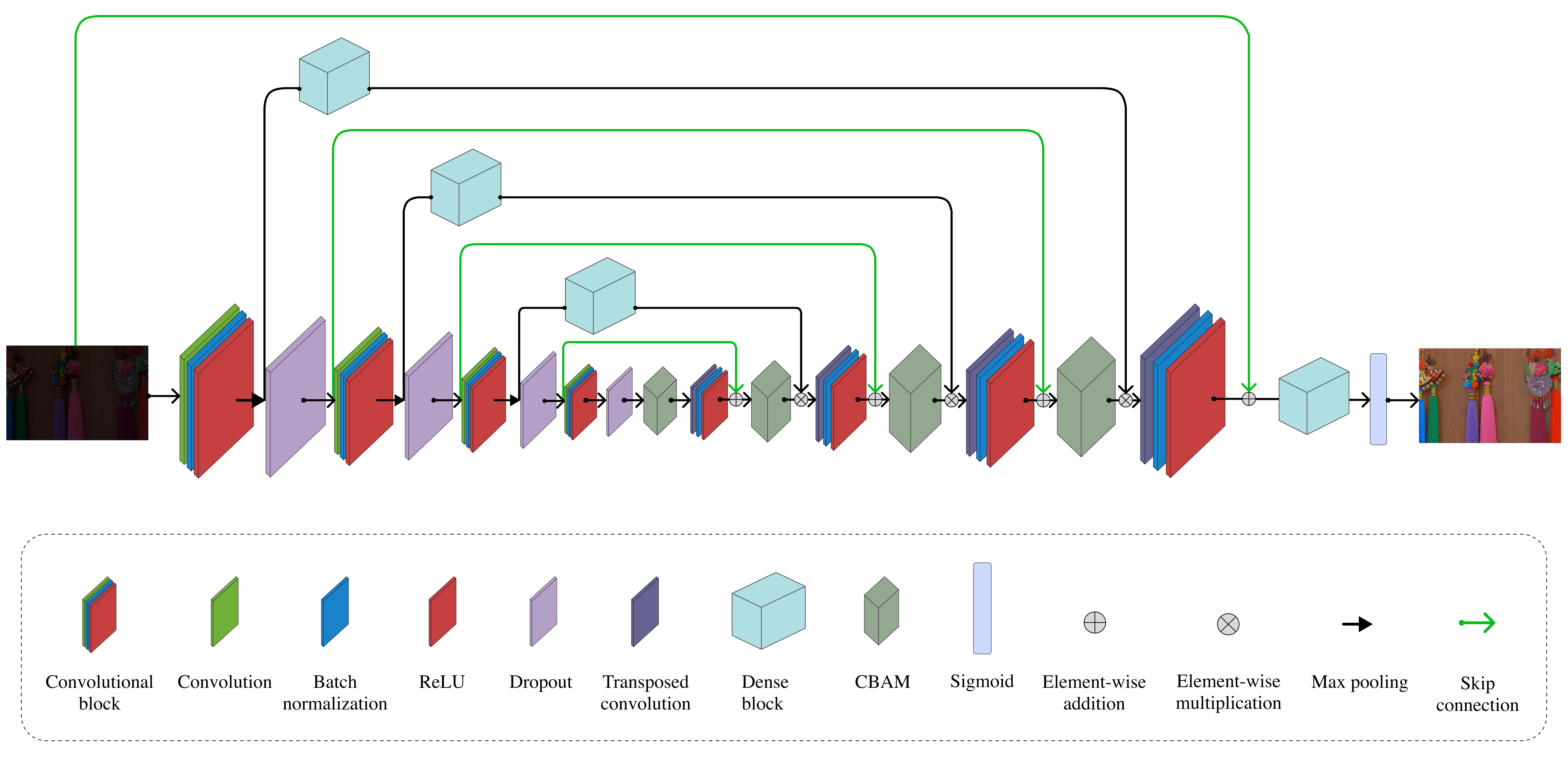}
    \caption{The overall structure of the proposed CDAN. The model integrates autoencoder-based architecture with convolutional and dense blocks, attention modules, and skip connections, facilitating efficient information propagation and feature learning.}
    \label{fig:fig1}
\end{figure*}

We propose a model architecture combining convolutional blocks, dense blocks, attention modules, and skip connections to enhance feature maps. In the encoding phase, convolutional blocks, max-pooling, and dropout \cite{Srivastava2014} layers process the input image $I_i \in \mathbb{R}^{H \times W \times 3}$ to yield feature maps of size $\frac{H}{8} \times \frac{W}{8} \times 512$. Three branched dense blocks capture intricate features, and four strategic skip connections link the encoder and decoder. For feature enhancement, we employ the Convolutional Block Attention Module (CBAM) \cite{Woo2018} within the bottleneck, prioritizing important spatial regions and channel-wise information. In the decoding phase, four transposed convolutional layers with batch normalization \cite{Ioffe2015} and rectified linear unit (ReLU) activation \cite{nair2010rectified} enable up-sampling. Skip connections and dense block branches are strategically incorporated, with CBAM modules persisting in the decoder. A final 4-layer dense block refines features before generating the ultimate output. Figure \ref{fig:fig1} provides an overview of the model structure, and the subsequent sections elaborate on core sub-modules of the architecture with more details.

\subsubsection{Convolutional block}
The convolutional block submodule plays a fundamental role in our model's encoder architecture by enabling effective feature extraction. Comprising a convolutional layer, batch normalization, and ReLU activation, this submodule captures essential spatial patterns from input data. The convolutional layer convolves the input, batch normalization enhances training stability, and ReLU introduces non-linearity. These components synergistically contribute to feature enrichment. 

\subsubsection{Dense block}
The dense block submodule constitutes a pivotal component within our model, enhancing feature representation by facilitating intricate concatenation. Inspired by the DenseNet architecture \cite{huang2017densely}, this submodule employs a stack of convolutional layers, each utilizing batch normalization and ReLU activation, fostering information propagation across different network depths. We incorporate three 4-layer dense blocks within the encoder, each with a growth rate of 16. These dense blocks are then multiplied into various decoder depths, contributing to feature enrichment during the decoding process. Furthermore, a 4-layer dense block with a growth rate of 16 is positioned at the end of the decoder, enhancing the refinement of the final feature maps.

\subsubsection{Convolutional block attention module}
The Convolutional Block Attention Module (CBAM) \cite{Woo2018} emerges as a powerful asset in the pursuit of enriched feature representation. Operating sequentially, CBAM extracts dual attention maps from an intermediate feature map \( F \in \mathbb{R}^{C \times H \times W} \), including a 1D channel attention map \( M_c \in \mathbb{R}^{C \times 1 \times 1} \) and a 2D spatial attention map \( M_s \in \mathbb{R}^{1 \times H \times W} \). The process can be expressed as:
\begin{equation}
\begin{aligned}
    F' &= M_c(F) \odot F, \\
    F'' &= M_s(F') \odot F',
\end{aligned}
\end{equation}
where \( \odot \) denotes element-wise multiplication and attention values are distributed accordingly, extending across spatial dimensions for channel attention values and vice versa. The resulting \( F'' \) represents a feature map enriched with focused details \cite{Woo2018}.

\subsection{Loss function}
We train our model using a composite loss function that combines Mean Squared Error (MSE) loss with a perceptual loss inspired by the VGG19 \cite{simonyan2015deep} network. By integrating this perceptual loss, we aim to enhance the output's alignment with perceptual quality, mirroring human perception characteristics. The loss function is defined as follows:
\begin{equation}\label{composite_loss}
\mathcal{L}_{\text{composite}} = \mathcal{L}_{\text{MSE}} + \lambda\mathcal{L}_{\text{VGG}},
\end{equation}
where \(\mathcal{L}_{\text{MSE}}\) is the MSE loss calculated as:
\begin{equation}\label{eq:mse}
\mathcal{L}_{\text{MSE}} = \frac{1}{N} \sum_{i=1}^{N} \| \hat{I}_i - I_i \|_2^2,
\end{equation}
here, $\hat{I}_i$ indicates the model's output, $I_i$ represents the ground truth, and $\mathcal{L}_{\text{perceptual}}$ signifies the perceptual loss obtained from the VGG19 network's intermediate feature maps. The hyperparameter $\lambda$ controls the balance between the two components.

Mathematically, the perceptual loss is expressed as follows:
\begin{equation}
\mathcal{L}_{\text{VGG}} = \frac{1}{N} \sum_{i=1}^{N} \| VGG(\hat{I}_i) - VGG(I_i) \|_2^2,
\end{equation}
where $N$ represents the batch size, $I_i$ denotes the $i$-th input image, and $VGG(\hat{I}_i)$ and $VGG(I_i)$ refer to the feature maps extracted from the output of our proposed model ($\hat{I}_i$) and the corresponding ground truth image ($I_i$), respectively. 

The model is trained to minimize this composite loss using the Adam optimizer \cite{kingma2017adam}, resulting in outputs that adhere to pixel-wise fidelity and encompass perceptual similarity to the ground truth.

\subsection{Post-processing}
Post-processing techniques are crucial in enhancing the quality of low-light images produced by deep learning models. Through post-processing, it is possible to improve quantitative metrics and enhance the overall image quality. Our image post-processing methodology draws inspiration from the interpolation and extrapolation approach introduced by Haeberli et al. \cite{haeberli1994image}, implemented in the Pillow (PIL Fork) library. This technique controls color enhancement operations such as brightness, contrast, saturation, tint, and sharpness with one formula, separately or simultaneously.

\subsubsection{Color enhancement}
The color enhancement process involves blending the original image with its grayscale counterpart, as defined by the equation below:
\begin{equation}\label{eq:post-processing}
\text{Enhanced Image} = \text{Image}_1(1 - \alpha) + \text{Image}_2\alpha,
\end{equation}
where $\text{Image}_1$ represents the original image, and $\text{Image}_2$ corresponds to its grayscale version, both having the same dimensions. The interpolation factor, denoted as $\alpha$, controls the degree of influence of each image in the final result. By varying the value of $\alpha$, we can achieve different levels of color enhancement.

\subsubsection{Contrast enhancement}
Our contrast manipulation approach uses a constant gray image with average image luminance. Through interpolation, contrast is tempered, while extrapolation amplifies it. Negative $\alpha$ values yield inverted images with varying contrast. Notably, average image luminance remains consistent.

\section{Experimental results}
This section presents a comprehensive evaluation of the proposed CDAN model's performance. Our assessment covers quantitative and qualitative aspects to demonstrate the model's robustness. We compared our model's effectiveness to state-of-the-art methods for enhancing low-light images, utilizing four benchmark datasets: LOL \cite{Wei2019}, ExDark \cite{LOH2019}, DICM \cite{Lee2013}, and VV \cite{vvdataset}. Additionally, we carefully examined our model's structure by conducting ablation studies to understand the specific effects of each component.

\subsection{Dataset}
In this research, the LOw-Light (LOL) dataset \cite{Wei2019} was utilized as the primary dataset for training our model. The LOL dataset is a well-known benchmark dataset specifically curated for low-light image processing tasks. It consists of a total of 500 image pairs with a size of $600\times400\times3$ pixels, which are further divided into 485 training pairs and 15 testing pairs. Each pair in the dataset comprises a low-light image and its corresponding high-quality reference image. This dataset serves as an excellent resource for training and evaluating deep learning models aimed at enhancing low-light images.
  
Besides LOL, we expanded our evaluation to include three additional benchmark datasets, assessing our model's performance and robustness across various lighting conditions:

\textbf{ExDark.} The ExDark \cite{LOH2019} is a specialized benchmark tailored for object detection and recognition tasks involving extremely low-light conditions. ExDark contains an extensive collection of 7,363 low-light images captured in challenging lighting conditions.

\textbf{DICM.} The DICM \cite{Lee2013}, a curated dataset dedicated to low-light enhancement, comprises 69 images captured using commercial digital cameras. Spanning diverse scenarios, it encompasses both indoor and outdoor settings, effectively encapsulating an array of lighting conditions experienced throughout the day and night.

\textbf{VV.} The VV \cite{vvdataset} presents a rare challenge in local exposure correction and enhancement. Its images, containing both correctly exposed and severely under/overexposed sections, push algorithms to enhance while avoiding artifacts and maintaining local contrast selectively.

\subsection{Experimental setup}
For our experiments, we used the PyTorch framework to develop our model. Training operations were performed on an Nvidia GeForce RTX 3090 GPU, utilizing its computational power to perform deep learning computations efficiently. The Adam optimizer \cite{kingma2017adam} was used during the training process to facilitate model optimization. As a preprocessing step, the images were uniformly resized to dimensions of $200\times200\times3$ for training, ensuring consistency and optimal input size.

\subsection{Evaluation metrics}
In the domain of low-light image enhancement, a comprehensive evaluation of image performance entails the utilization of both quantitative and qualitative assessments. Quantitative assessment involves the application of objective measurements to provide numerical evaluations of the images. Well-established metrics such as Peak Signal-to-Noise Ratio (PSNR) \cite{Wang2004}, Structural Similarity Index Metric (SSIM) \cite{Wang2004}, and Learned Perceptual Image Patch Similarity (LPIPS) \cite{Zhang2018} are commonly employed to quantitatively compare the enhanced images against their corresponding ground truth reference images. These metrics enable us to obtain objective insights into crucial factors such as image fidelity, noise reduction, and overall similarity to the ground truth.

The PSNR measures the ratio between the maximum possible power of a signal and the power of the noise present in the signal. It is often used to evaluate image quality. The formula for PSNR is given by:
\begin{equation}
\text{PSNR (dB)} = 10 \cdot \log_{10}\left(\frac{{\text{MAX}_I^2}}{{\text{MSE}}}\right),
\end{equation}
where \(MAX_I\) represents the maximum pixel value of the image (e.g., 255 for an 8-bit image) and MSE is the mean squared error (defined by Equation \ref{eq:mse}) between the enhanced image and the ground truth reference image.

The SSIM is a metric that measures the similarity between two images. It considers the images' luminance, contrast, and structure. The formula for SSIM is given by:
\begin{equation}
\text{SSIM}(x, y) = \frac{{(2\mu_x\mu_y + C_1)(2\sigma_{xy} + C_2)}}{{(\mu_x^2 + \mu_y^2 + C_1)(\sigma_x^2 + \sigma_y^2 + C_2)}},
\end{equation}
where \(x\) and \(y\) represent the enhanced and ground truth reference images, respectively. \(\mu_x\) and \(\mu_y\) are the average values of \(x\) and \(y\), \(\sigma_x^2\) and \(\sigma_y^2\) are their variances, and \(\sigma_{xy}\) is the covariance of \(x\) and \(y\). \(C_1\) and \(C_2\) are small constants used to stabilize the division \cite{Wang2004}.

LPIPS is a perceptual similarity metric that measures the similarity between two images based on human perception. It considers color, texture, and structure to capture the perceptual difference between images. LPIPS is computed by comparing image patches extracted from the two images using a deep neural network. The lower the LPIPS score, the more similar the images are perceived to be.

\subsection{Hyperparameter tuning}
Hyperparameter tuning is critical for developing effective deep learning models for low-light image enhancement. The selection of appropriate hyperparameters significantly impacts the quality of the enhanced images produced by the model. In this study, we employed the Optuna library with the Tree-structured Parzen Estimator (TPE) sampler to search for optimal hyperparameter values efficiently. The TPE sampler is a Bayesian optimization algorithm that adaptively selects points for evaluation based on the previous good results. 

After concluding the hyperparameter tuning process, we identified the most effective values for the hyperparameters. These optimal values include 80 epochs, a batch size of 16, a learning rate of 0.001, a $\lambda$ factor of 0.25 for the loss function, as well as 20 VGG feature maps for the loss function. In the post-processing phase, we found that setting $\alpha$ factors of 1.12 for contrast enhancement and 1.35 for color enhancement led to significant improvements.

\subsection{Experiments on the LOL dataset}
To validate the performance of the proposed model, we conducted extensive experiments on the paired testing set of the LOL dataset, consisting of 15 images. We conducted a comprehensive evaluation of the data both quantitatively and qualitatively. To ensure consistent evaluation conditions, all testing images were standardized to a fixed size of $600\times400$ in height and width. For the majority of our comparative analysis, we utilized the research findings and online platform provided by Li et al. \cite{Li2022survey}. Our experiments followed the settings outlined in their comprehensive survey, which involved using an AlexNet-based model to calculate perceptual similarity for the LPIPS metric.

\subsubsection{Quantitative evaluation}
In this section, we perform a quantitative comparison between our proposed model and several state-of-the-art methods, namely LLNET \cite{LORE2017}, LightenNet \cite{LI2018}, MBLLEN \cite{Lv2018MBLLEN}, Retinex-Net \cite{Wei2018Retinex}, KinD \cite{Zhang2019}, Kind++ \cite{Zhang2021}, TBEFN \cite{lu2020tbefn}, DSLR \cite{Lim2021}, LAU-Net \cite{LIM2023}, DRBN \cite{Yang2021}, EnlightenGAN \cite{jiang2021}, ExCNet \cite{Zhang2019Zero-shot}, Zero-DCE \cite{Guo2020}, and RRDNet \cite{Zhu2020}. It's important to note that we extracted the results of all these methods, except for LAU-Net \cite{LIM2023}, from \cite{Li2022survey}. The comparison is based on three key metrics: average PSNR, SSIM, and LPIPS. These comparative results are presented in Table \ref{tab:comparison}.

It is clear from Table \ref{tab:comparison} that our proposed model exhibits superior performance across different learning methods. CDAN achieves the highest SSIM score of 0.816 across all evaluated methods. This is a critical achievement, as SSIM is a perceptual metric that quantifies image quality degradation resulting from data loss during compression or noise interference, which indicates that our model excels in preserving the structural and textural details of the original image. The strong PSNR score of 20.102 and low LPIPS score of 0.167 further validate its precision and perceptual quality, reflecting its capacity to enhance low-light images with remarkable fidelity.

Among the supervised methods, LightenNet \cite{LI2018} appears as the least effective performer, yielding an average PSNR of 10.301 and an average SSIM of 0.402. These figures underscore the challenges of retaining fine image details and maintaining perceptual similarity. LLNET \cite{LORE2017} demonstrates decent results with an average PSNR of 17.959. However, it utilizes a substantial parameter count of 17.908 million \cite{Li2022survey}. MBLLEN \cite{Lv2018MBLLEN}, a multi-branch network, presents an encouraging average PSNR of 17.902 and SSIM of 0.715, suggesting compelling perceptual similarity. Retinex-Net \cite{Wei2018Retinex} performs well regarding average PSNR (16.774) but lags in SSIM and LPIPS values, hinting at potential issues in detail preservation and perceptual quality. Our model achieves an impressive LPIPS score of 0.167, claiming the best position, while KinD \cite{Zhang2019} follows closely with 0.175. KinD++ \cite{Zhang2021} secures a strong LPIPS score of 0.198. These results underscore the effectiveness of our model and these methodologies in preserving visual quality, enhancing details, and improving overall image appearance while addressing low-light conditions. LAU-Net \cite{LIM2023}, trained on the LOL dataset, distinguishes itself with the highest average PSNR (21.513) and robust SSIM performance (0.805), thereby illuminating the transformative effects of harnessing U-net-based architectures alongside the application of attention modules in the domain of low-light image enhancement tasks.

In the case of the semi-supervised approach, DRBN \cite{Yang2021} demonstrates outcomes characterized by a moderate average PSNR (15.125) and SSIM (0.472), pointing to an image enhancement quality that falls short of being satisfactory. In contrast, EnlightenGAN \cite{jiang2021}, an unsupervised technique, offers a more well-rounded performance, achieving an average PSNR of 17.483 and an SSIM score of 0.322. This makes EnlightenGAN a dependable contender in low-light image enhancement methods.

Zero-shot learning methods have the distinct advantage of not requiring paired data for training. This is a considerable benefit as collecting paired data is resource-intensive and often impractical. Exemplifying this, ExCNet \cite{Zhang2019Zero-shot} exhibits a relatively balanced performance across all three metrics, with an average PSNR of 15.783, SSIM score of 0.515, and LPIPS score of 0.373. Comparatively, Zero-DCE \cite{Guo2020} outperforms ExCNet \cite{Zhang2019Zero-shot} in maintaining image details with its higher SSIM score of 0.589, yet ExCNet achieves superior scores in PSNR and LPIPS. Meanwhile, RRDNet \cite{Zhu2020} exhibits the weakest zero-shot performance, with an average PSNR of 11.392, SSIM score of 0.468, and LPIPS score of 0.361. These zero-shot methods, while requiring quantitative improvement, offer a unique perspective. Their capacity to learn from unpaired data showcases a departure from traditional supervised learning, unveiling a promising avenue in low-light image enhancement research. However, as highlighted in Table \ref{tab:comparison}, it is notable that supervised learning-based methods currently demonstrate superior performance and more pleasing perceptual outcomes for low-light image enhancement.

\begin{table*}
    \centering
    \caption{Performance comparison with deep learning-based state-of-the-art approaches on the LOL dataset, based on average PSNR, average SSIM, and average LPIPS metrics, categorized by learning methods. Higher PSNR and SSIM imply better performance, while lower LPIPS signify improved results. The best results are in \textcolor{red}{red} whereas the second-best results are in \textcolor{blue}{blue}.}\vspace{5pt}
    \label{tab:comparison}
    \begin{tabular}{cccccc}
    \toprule
    \textbf{Learning method} & \textbf{Method} & \textbf{Avg. PSNR} $\uparrow$ & \textbf{Avg. SSIM} $\uparrow$ & \textbf{Avg. LPIPS} $\downarrow$\\
    \midrule
    \multirow{11}{*}{Supervised}
    & LLNET \cite{LORE2017} & 17.959 & 0.713 & 0.360\\
    & LightenNet \cite{LI2018} & 10.301 & 0.402 & 0.394\\
    & MBLLEN \cite{Lv2018MBLLEN} & 17.902 & 0.715 & 0.247\\
    & Retinex-Net \cite{Wei2018Retinex} & 16.774 & 0.462 & 0.474\\
    & KinD \cite{Zhang2019} &17.648 & 0.779 & \textcolor{blue}{0.175}\\
    & Kind++ \cite{Zhang2021} & 17.752 & 0.760 & 0.198\\
    & TBEFN \cite{lu2020tbefn} & 17.351 & 0.786 & 0.210\\
    & DSLR \cite{Lim2021} & 15.050 & 0.597 & 0.337\\
    & LAU-Net \cite{LIM2023} & \textcolor{red}{21.513} & \textcolor{blue}{0.805} & 0.273\\
    \midrule
    \multirow{1}{*}{Semi-supervised} & DRBN \cite{Yang2021} & 15.125 & 0.472 & 0.316\\
    \midrule
    \multirow{1}{*}{Unsupervised} & EnlightenGAN \cite{jiang2021} & 17.483 & 0.677 & 0.322\\
    \midrule
    \multirow{2}{*}{Zero-shot}
    & ExCNet \cite{Zhang2019Zero-shot} & 15.783 & 0.515 & 0.373\\
    & Zero-DCE \cite{Guo2020} & 14.861 & 0.589 & 0.335\\
    & RRDNet \cite{Zhu2020} & 11.392 & 0.468 & 0.361\\
    \midrule
    & Proposed (CDAN) & \textcolor{blue}{20.102} & \textcolor{red}{0.816} & \textcolor{red}{0.167}\\
    \bottomrule
    \end{tabular}
\end{table*}

\subsubsection{Qualitative evaluation}
\begin{figure}
    \centering
    \includegraphics[width=\linewidth]{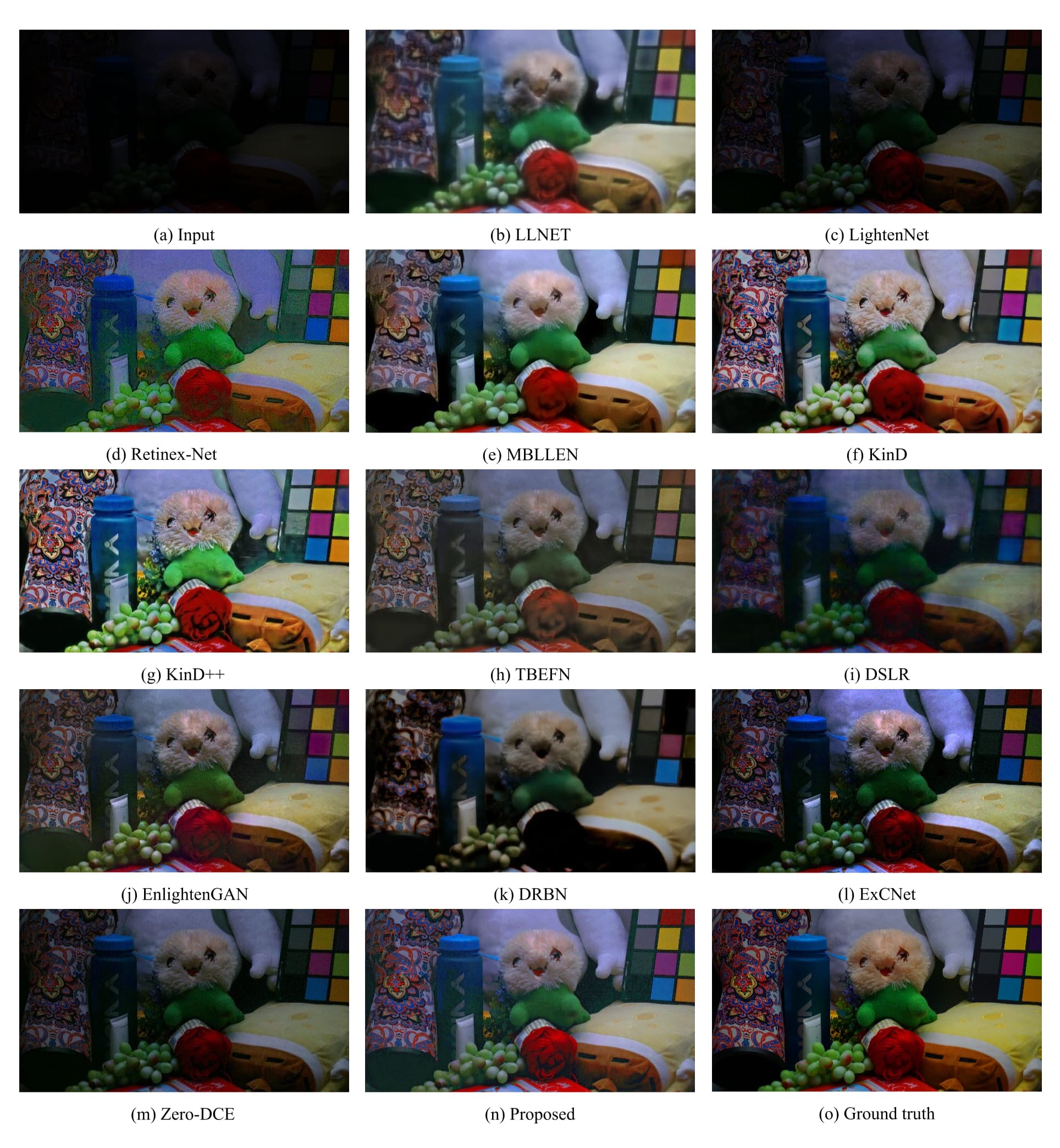}
    \caption{Comparing deep learning-based state-of-the-art approaches on a low-light image from the LOL test dataset.}
    \label{fig:fig2}
\end{figure}

In this section, we qualitatively compare our proposed approach and existing deep learning-based state-of-the-art methods. To illustrate the efficacy and robustness of our model, we have chosen three images from the test set of the LOL dataset. Illustrated in Figure \ref{fig:fig2}, our model has demonstrated exceptional performance in enhancing image contrast while accurately restoring colors and intricate details. In contrast, several methods, such as LightenNet \cite{LI2018}, DSLR \cite{Lim2021}, ExCNet \cite{Zhang2019Zero-shot}, and Zero-DCE \cite{Zhang2019Zero-shot}, still exhibit under-exposure issues, and others struggle to restore complete image details and accurate color representation, often accompanied by artifacts and color distortion.

\begin{figure}
    \centering
    \includegraphics[width=\linewidth]{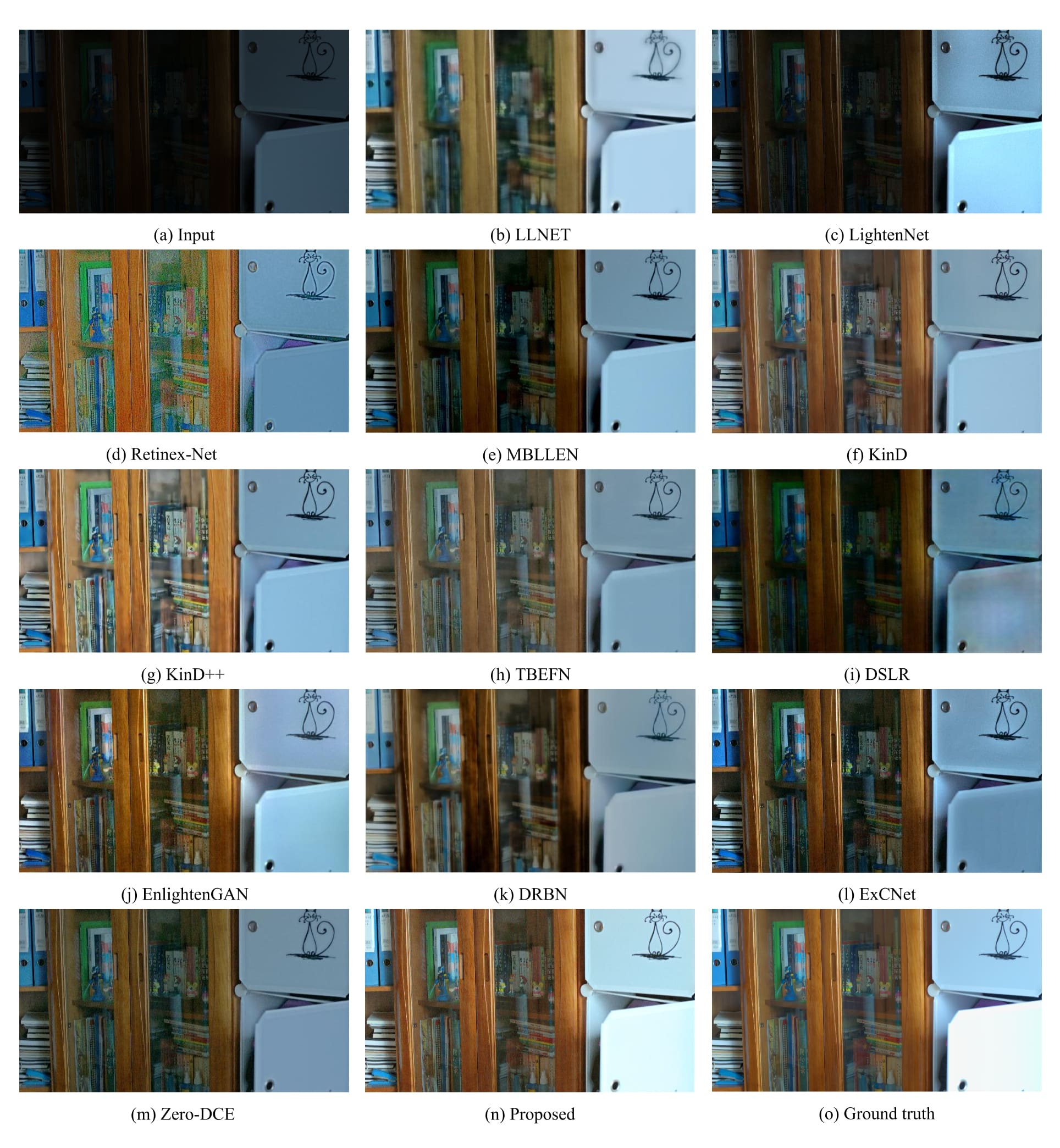}
    \caption{Comparing deep learning-based state-of-the-art approaches on a low-light image from the LOL test dataset.}
    \label{fig:fig3}
\end{figure}

Illustrated in Figure \ref{fig:fig3}, LightenNet \cite{LI2018} continues to grapple with under-exposure issues, RetinexNet \cite{Wei2018Retinex} exhibits color distortion, and several other methods showcase color bias or inability to restore light to the image adequately. Notably, our proposed approach enhances the image in a highly natural manner, effectively mitigating both under and over-exposure issues while maintaining uniform lighting. KinD \cite{Zhang2019}, TBEFN \cite{lu2020tbefn}, and EnlightenGAN \cite{jiang2021} also achieve comparable results.

\begin{figure}
    \centering
    \includegraphics[width=\linewidth]{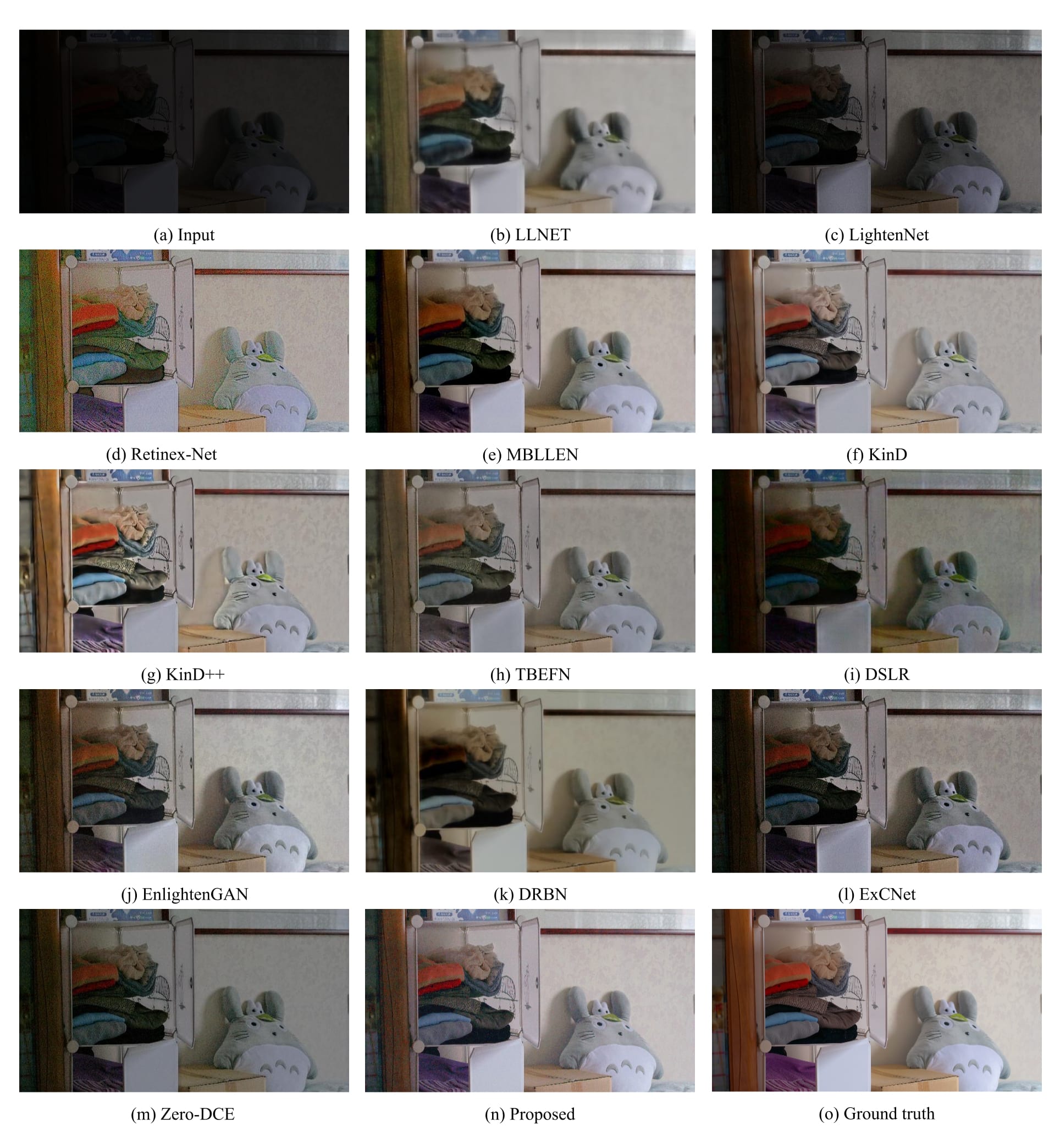}
    \caption{Comparing deep learning-based state-of-the-art approaches on a low-light image from the LOL test dataset.}
    \label{fig:fig4}
\end{figure}

As depicted in Figure \ref{fig:fig4}, LLNET \cite{LORE2017} exhibits blurriness in its result, while LightenNet \cite{LI2018} still suffers from under-exposure. The result produced by Retinex-Net \cite{Wei2018Retinex} is negatively impacted by significant noise, and KinD++ \cite{Zhang2021} fails to achieve a natural outcome. Like LLNET, DRBN \cite{Yang2021} also experiences blurriness and difficulty restoring colors accurately. DSLR \cite{Lim2021}, ExCNet \cite{Zhang2019Zero-shot}, and Zero-DCE \cite{Zhang2019Zero-shot} must be improved by adequate lighting, leading to suboptimal results. However, MBLLEN \cite{Lv2018MBLLEN}, KinD \cite{Zhang2019}, and EnlightenGAN \cite{jiang2021} present promising results regarding image enhancement. Impressively, our proposed approach excels in restoring diverse colors naturally and ensuring well-lit, high-quality results across the image.

Overall, as demonstrated in the presented figures, our CDAN model showcases exceptional efficacy in enhancing low-light images, yielding natural and perceptually pleasing results.

\subsection{Experiments on varied lighting conditions}
In deep learning, the robustness of models to unseen data, particularly in challenging low-light image enhancement tasks, is of paramount importance. This section delves into an in-depth exploration of our proposed model's performance across varying lighting scenarios. Through rigorous experimentation, we compare our model against other leading deep learning approaches using diverse datasets. This analysis not only highlights our model's strengths but also critically evaluates its limitations, providing insights into areas that require further refinement to bolster its adaptability across different lighting conditions.

\begin{figure}
    \centering
    \includegraphics[width=\linewidth]{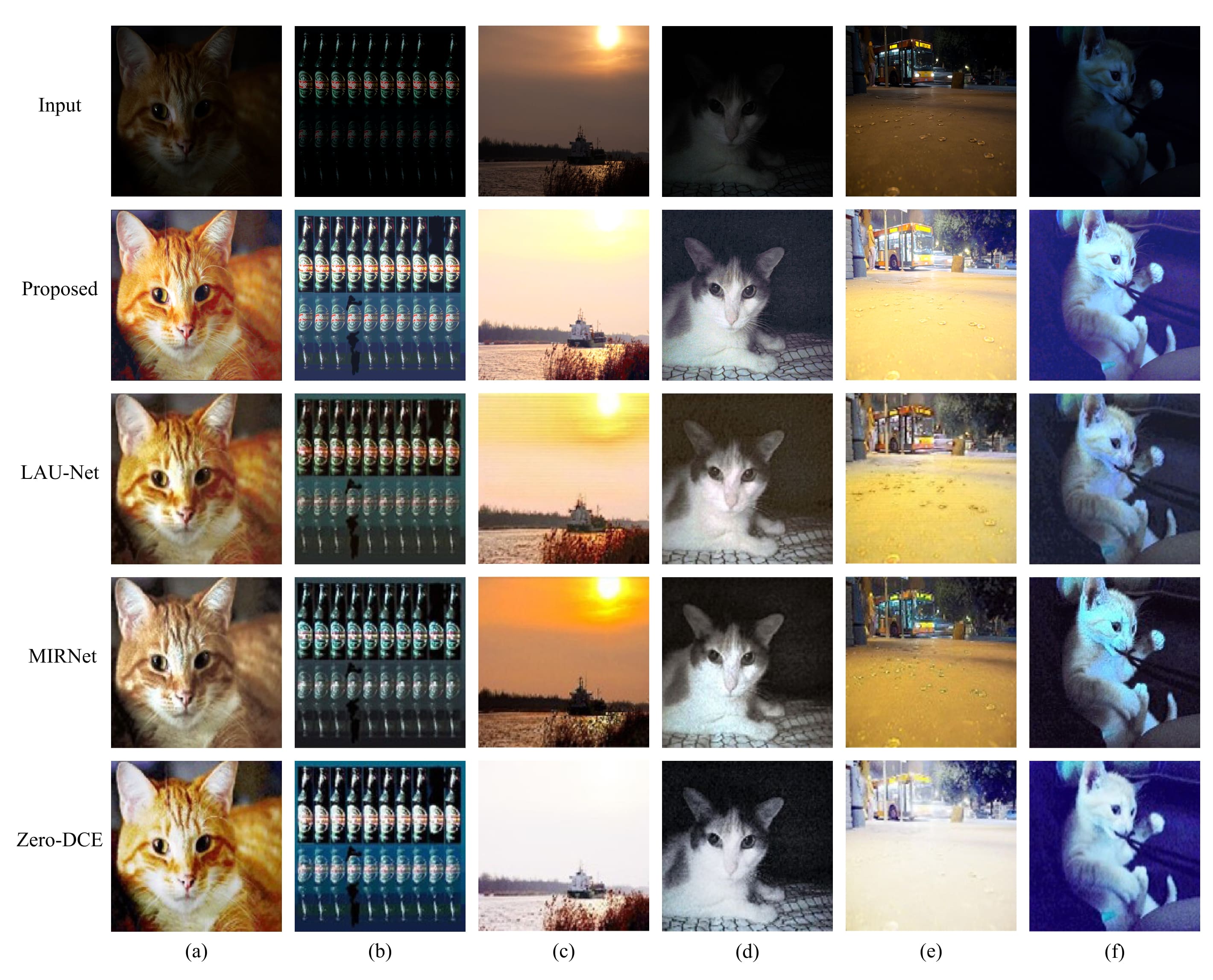}
    \caption{Comparing deep learning-based state-of-the-art approaches on six low-light images from the ExDark dataset.}
    \label{fig:fig5}
\end{figure}

Figure \ref{fig:fig5} compares our CDAN model with three other high-performing deep learning-based low-light enhancers, including LAU-Net \cite{LIM2023}, MIRNet \cite{Zamir2020}, and Zero-DCE \cite{Guo2020}, using the ExDark dataset as the testing ground. Our CDAN model delivers remarkable advancements in enhancing low-light images while preserving their intricate details. As illustrated in Figure \ref{fig:fig5}(a), our model can significantly enhance the colors, textures, and details within the cat's fur patterns. Based on Figure \ref{fig:fig5}(b), it is evident that our model surpasses other methods in enhancing the bottle labels. Additionally, the output produced by our model exhibits superior visual quality. In Figure \ref{fig:fig5}(c), LUA-Net \cite{LIM2023}, having the highest PSNR score on the LOL dataset, introduces visible artifacts in the sky. MIRNet \cite{Zamir2020} struggles with under-exposure, while Zero-DCE \cite{Guo2020} exaggerates over-exposure, obliterating the sunset and turning the sky uniformly white. Our model adeptly enhances the ship's colors at sea while retaining the hues and details of the sky with a small quantity of over-enhancement of the sun. All models demonstrated commendable performance in Figure \ref{fig:fig5}(d), achieving acceptable levels of visual detail. Figure \ref{fig:fig5}(e) again exposes that Zero-DCE \cite{Guo2023} tends to over-expose, while LAU-Net \cite{LIM2023} effectively restores details such as the footprints compared to other models. LAU-Net \cite{LIM2023} stands out in Figure \ref{fig:fig5}(f) with exceptional performance.

\begin{figure}
    \centering
    \includegraphics[width=\linewidth]{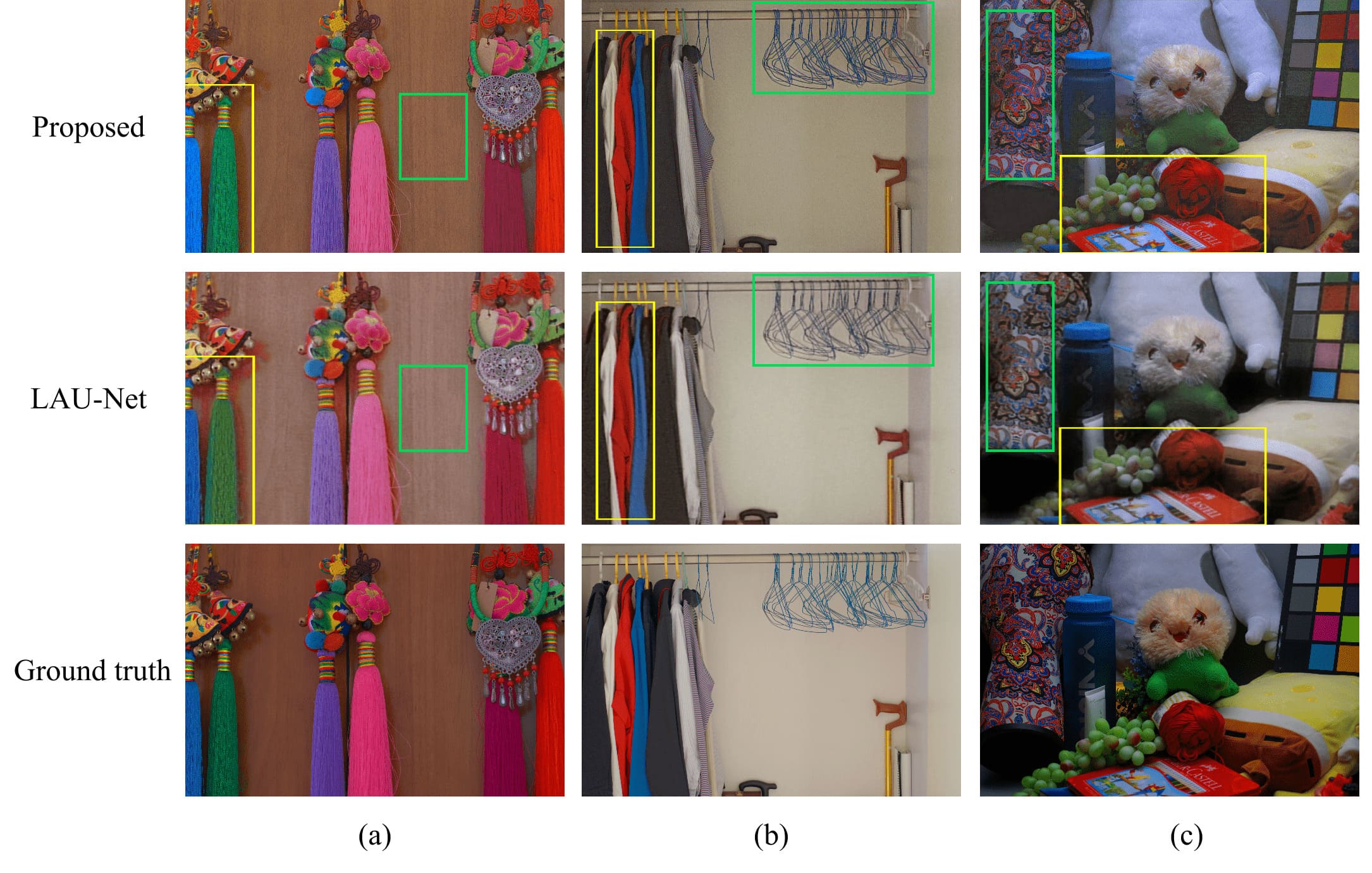}
    \caption{Comparing the performance of our proposed model and LAU-Net \cite{LIM2023} on images from the LOL dataset. The yellow boxes highlight texture preservation and detail restoration, while the green boxes emphasize color refinement.}
    \label{fig:fig6}
\end{figure}

Figure \ref{fig:fig6} compares the performance of our proposed model and LAU-Net \cite{LIM2023} on images from the LOL dataset. Although the quantitative results of both models are close, Figure \ref{fig:fig6} shows that our model outperforms LAU-Net \cite{LIM2023} in qualitative aspects such as color refinement, texture preservation, and detail restoration.

\begin{figure}
    \centering
    \includegraphics[width=\linewidth]{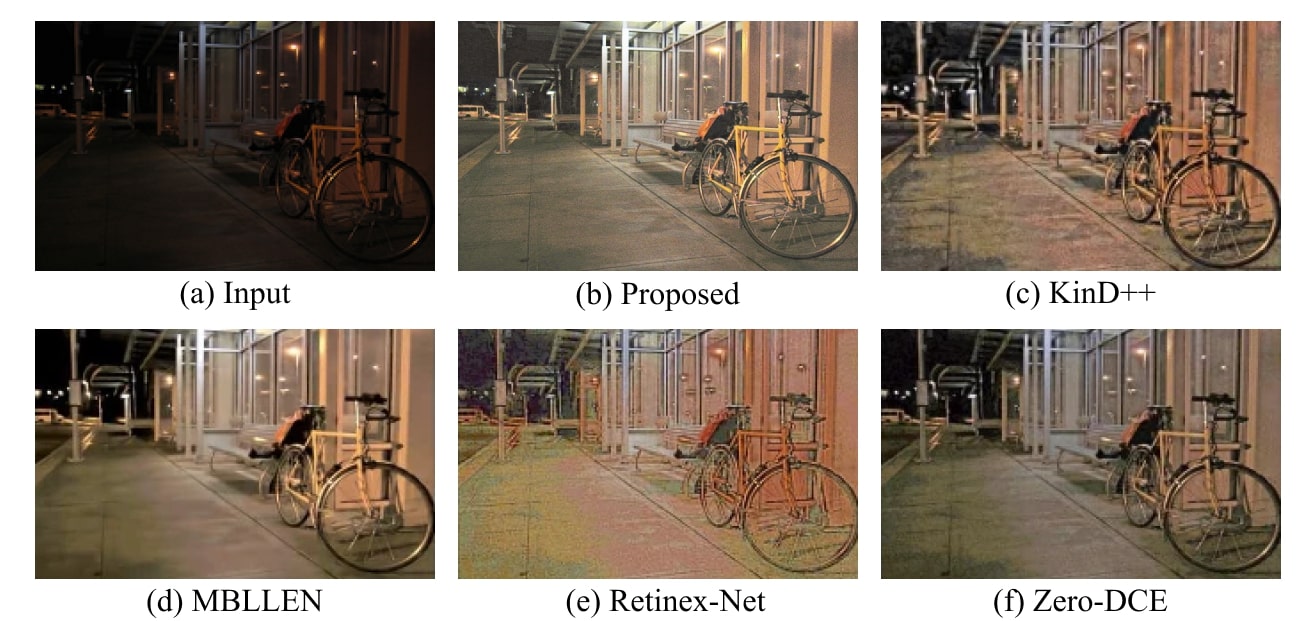}
    \caption{Comparing deep learning-based state-of-the-art approaches on a low-light image from the ExDark dataset.}
    \label{fig:fig7}
\end{figure}

Figure \ref{fig:fig7} captures a challenging low-light scenario extracted from the ExDark dataset, marked by diverse illumination areas. Retinex-Net's \cite{Wei2018Retinex} performance falls short among the evaluated models, exhibiting color distortion, noise, and a notable loss of information. Similarly, KinD++ \cite{Zhang2021} faces challenges in generating a coherent scene marked by noticeable artifacts and noise. Zero-DCE \cite{Guo2020} manages to create a natural-looking well-lit image, yet it needs to improve in the refinement of all colors. While MBLLEN \cite{Lv2018MBLLEN} achieves a visually appealing outcome, a subtle blurriness is evident on the pavement. Notably, our CDAN model excels, achieving remarkable balance by providing sufficient lighting while expertly preserving the scene's authentic colors.

\begin{figure}
    \centering
    \includegraphics[width=\linewidth]{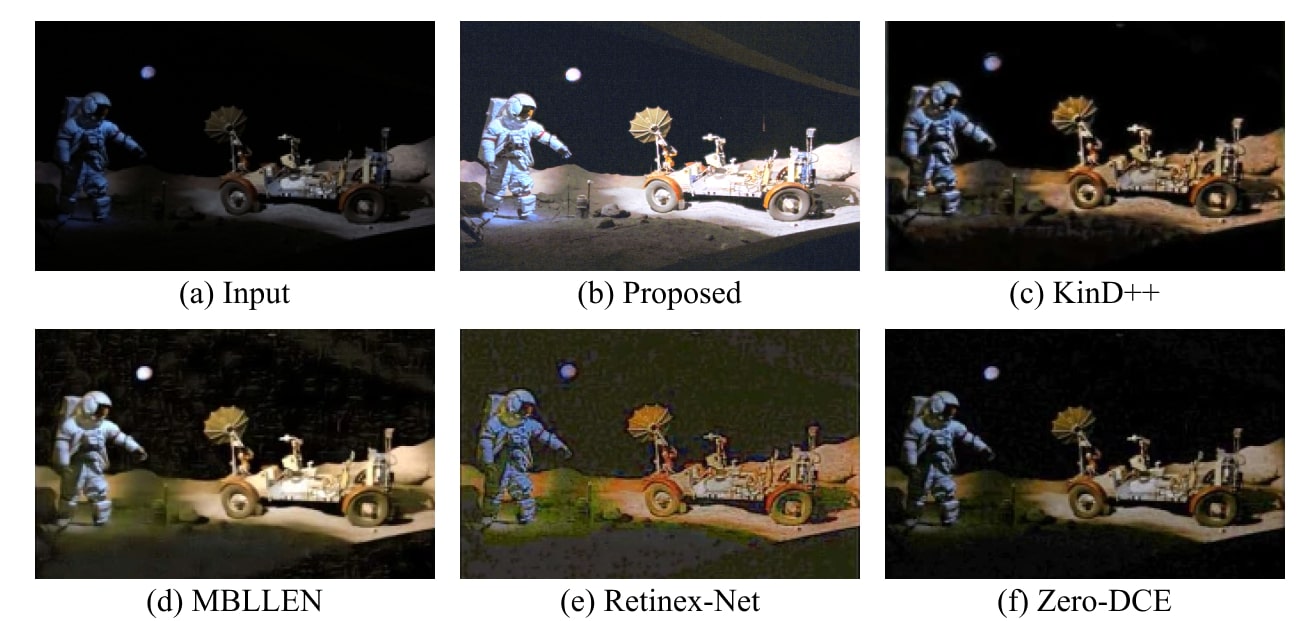}
    \caption{Comparing deep learning-based state-of-the-art approaches on a low-light image from the DICM dataset.}
    \label{fig:fig8}
\end{figure}

Figure \ref{fig:fig8} showcases the exceptional performance of our CDAN model, outperforming its competitive counterparts. The astronaut in the image generated by our model benefits from excellent lighting, with his spacesuit's colors well-preserved and vividly presented. The image has a pleasant contrast that distinguishes it from other models. Even minute details, such as the tiny stones on the astronaut's standing surface, are discernible. However, it is worth noting that the machinery component is slightly over-exposed, creating excessive brightness. In comparison, Retinex-Net \cite{Wei2018Retinex} contends with pronounced noise and fails to achieve a perceptually pleasing enhancement. Zero-DCE \cite{Guo2020} grapples with under-exposure, leading to a less-than-optimal enhancement outcome. KinD++ \cite{Zhang2021} shares similarities with Zero-DCE \cite{Guo2020} but improves performance by enhancing the machine's area within the image. Furthermore, MBLLEN demonstrates promising results, albeit with noticeable artifacts in the black regions of the image.

\begin{figure}
    \centering
    \includegraphics[width=\linewidth]{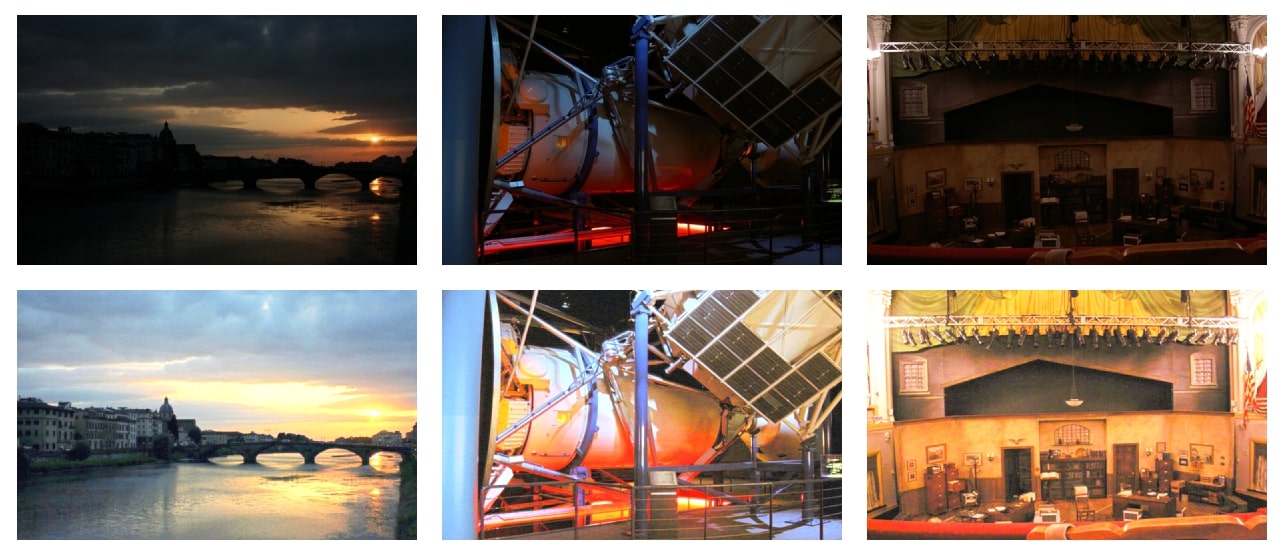}
    \caption{A comparison between low-light input images (top row) from the DICM dataset and enhanced outputs (bottom row) generated by our proposed model.}
    \label{fig:fig9}
\end{figure}

To further assess the robustness of our CDAN model, we have demonstrated enhanced outputs across a diverse range of low-light images, encompassing various lighting conditions indoors and outdoors throughout both day and night, as shown in Figure \ref{fig:fig9} using the DICM dataset. The figure vividly illustrates the model's capacity to illuminate intricate details and remarkably restore colors. It becomes evident that our model maintains robustness across many different lighting scenarios and environments.

\begin{figure}
    \centering
    \includegraphics[width=\linewidth]{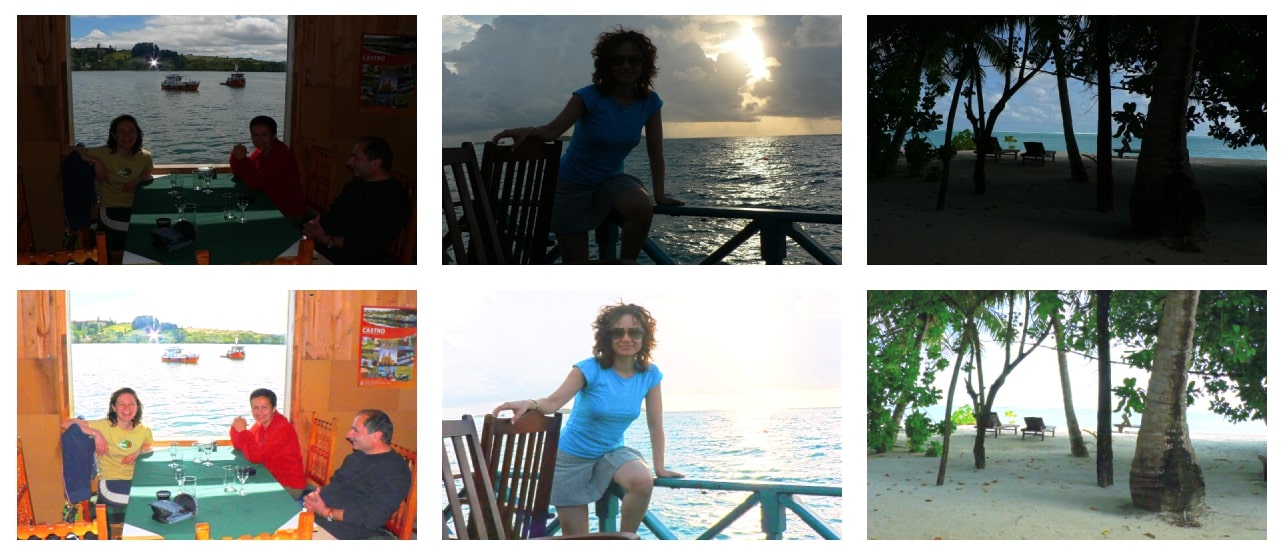}
    \caption{A comparison between low-light input images (top row) from the VV dataset and enhanced outputs (bottom row) generated by our proposed model. Our model displays a vulnerability in low-light scenarios, where it tends to over-enhance well-lit or over-exposed areas, such as sunlit regions.}
    \label{fig:fig10}
\end{figure}

Moving on to Figure \ref{fig:fig10}, we present enhanced images sourced from the VV dataset, spotlighting the challenges that underscore areas of improvement for our model in future iterations. These images present complex lighting variations within the same frame, notably featuring over-exposed and under-exposed regions. The challenge lies in distinguishing between properly exposed and degraded sections. Sometimes, our model tends to over-enhance certain areas, leading to over-exposed regions within an image. These limitations are evident in the enhanced images showcased in Figure \ref{fig:fig10}, where elements like the sun, sky, and highly sunlit objects reveal instances where our model struggles to maintain the desired balance.

We believe that the limitations of our model are primarily due to the characteristics of the dataset it was trained on. The LOL dataset is relatively small, synthesized, and consists of uniformly illuminated indoor scenes. We utilized it because it allows for low-cost training in a shorter time and is a paired dataset. Although our model demonstrates great capability in our comprehensive experiments, its performance could be improved by training on a more diverse and larger dataset that includes complex lighting scenarios. Future research should focus on using such datasets to enhance the model's robustness in handling scenes with high contrast in lighting.

In summary, these experiments conclusively establish our model as a state-of-the-art solution that excels or maintains parity with advanced low-light image enhancement models across a spectrum of lighting conditions. Its exceptional performance in extremely dark scenarios and scenes characterized by uniformly dim areas is particularly noteworthy. However, areas that include intensely illuminated regions pose challenges that need addressing.

\subsection{Ablation studies}
Our ablation studies comprehensively explore different components of our proposed CDAN model. This includes an assessment of different mechanisms and modules, along with an analysis of the impact of various loss functions on the performance of our model and the quality of the generated images, as well as a study on the effects of post-processing.

\begin{table*}
    \caption{Analyzing the impact of various model components on the LOL dataset via ablation studies, reported without post-processing.}
    \centering
    \begin{tabular}{ccccc}
    \toprule
    \textbf{No.} & \textbf{Components} & \textbf{Avg. PSNR} $\uparrow$ & \textbf{Avg. SSIM} $\uparrow$ & \textbf{Avg. LPIPS} $\downarrow$\\
    \midrule
    (a) & Basic autoencoder (convolutional blocks) & 14.551 & 0.594 & 0.393\\
    (b) & Basic autoencoder + Skip connections & 18.752 & 0.685 & 0.252\\
    (c) & Basic autoencoder + Skip connections + Attention & 18.372 & 0.782 & 0.212\\
    (d) & Basic autoencoder + Skip connections + Attention + Dense blocks & \textbf{19.569} & \textbf{0.820} & \textbf{0.164}\\
    \bottomrule
    \end{tabular}
    \label{tab:ablation_components}
\end{table*}

\subsubsection{Study on the effects of model components}
\begin{figure}
    \centering
    \includegraphics[width=\linewidth]{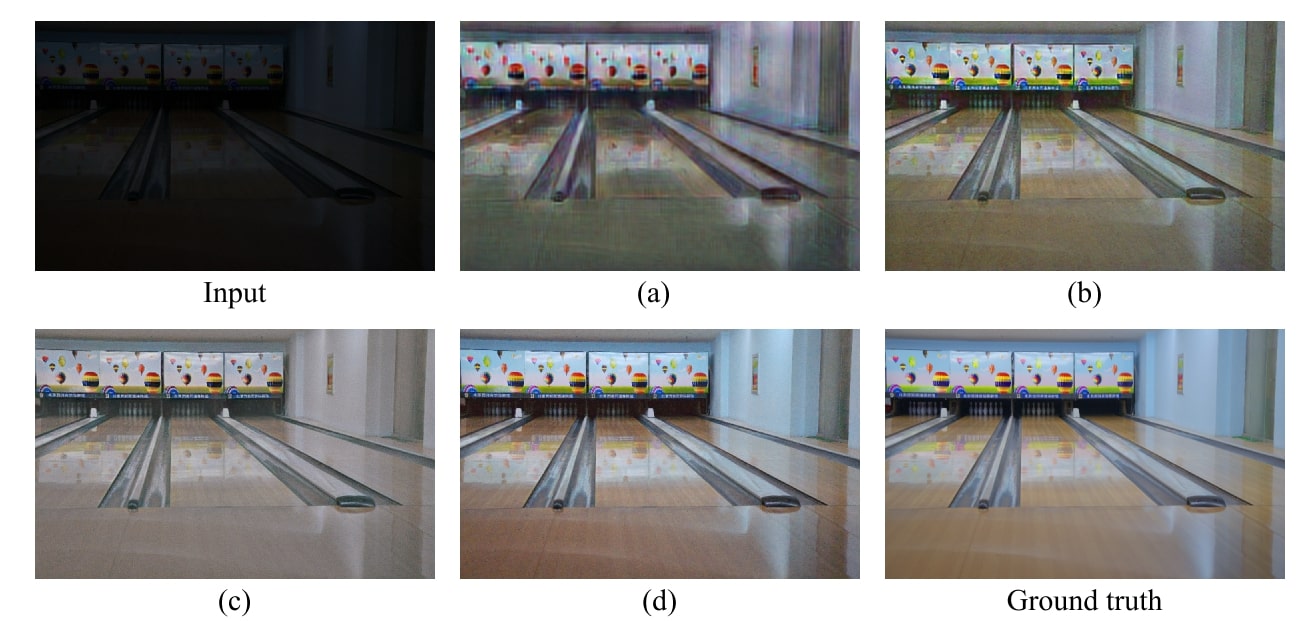}
    \caption{Enhanced images from the LOL dataset corresponding to the ablation studies in Table \ref{tab:ablation_components}, reported without post-processing. The image sequence shows the enhancement achieved through each architectural component.}
    \label{fig:fig11}
\end{figure}

In this section, we examine the effects of each mechanism and module within our proposed architecture on enhancing low-light images. The outcomes of our exploration are summarized in Table \ref{tab:ablation_components}, which contains the results of our detailed ablation study without any post-processing. Additionally, Figure \ref{fig:fig11} visually compares the improved LOL images based on the findings from Table \ref{tab:ablation_components}. Our goal is to understand how each module influences the network's efficiency and visual quality.

Initial implementation with just the basic autoencoder, comprising convolutional blocks, yields an average PSNR of 14.551, SSIM of 0.594, and LPIPS of 0.393. These results provide a baseline for comparing the improvements achieved by adding more sophisticated modules. The visual result of the basic autoencoder is represented in Figure \ref{fig:fig11}(a).

Adding skip connections to the basic autoencoder significantly improves all the performance metrics. The PSNR increases to 18.752, indicating a better reconstruction quality of the enhanced images. The SSIM improves to 0.685, suggesting that the enhanced images preserve more structural and textural details from the original image than the basic autoencoder. Meanwhile, the LPIPS decreases to 0.252, indicating that the perceptual difference between the original and enhanced images has lessened. These improvements, as shown in Figure \ref{fig:fig11}(b), demonstrate the value of skip connections in facilitating the information flow across the network and preserving low-level details.

Next, incorporating an attention mechanism, the Convolutional Block Attention Module (CBAM), alongside the basic autoencoder and skip connections results in a slight decrease in the PSNR to 18.372. However, the SSIM score substantially improves to 0.782, and the LPIPS score reduces further to 0.212. The decrease in PSNR but an increase in SSIM and reduction in LPIPS suggest that while the reconstructed image may slightly deviate from the original regarding global error (as represented by PSNR), it maintains better perceptual quality and structural details. This phenomenon underscores the limitation of using PSNR as the sole evaluation metric, as it does not fully capture the perceptual and structural improvements achieved by the CBAM attention mechanism. The attention mechanism helps the network focus on more relevant features, improving the visual quality of the enhanced images. The output of the proposed model after utilizing the attention mechanism is depicted in Figure \ref{fig:fig11}(c).

Finally, integrating dense blocks with the basic autoencoder, skip connections, and attention mechanism leads to the best overall performance. The model achieves the highest PSNR of 19.569, SSIM of 0.820, and the lowest LPIPS of 0.164. Dense blocks encourage feature reuse and improve gradient flow, enhancing the model's learning capacity. As a result, the quality of the enhanced image, as shown in Figure \ref{fig:fig11}(d), improves numerically and visually.

In conclusion, each additional component in our architecture incrementally improves our model's performance, except for a slight decrease in PSNR when introducing the attention mechanism. However, this decrease is accompanied by significant improvements in other metrics, indicating a better perceptual and structural quality of the enhanced images. Therefore, this ablation study underscores the importance of considering multiple metrics to evaluate the performance of low-light image enhancement models.

\subsubsection{Study on different loss functions}
\begin{figure}
    \centering
    \includegraphics[width=\linewidth]{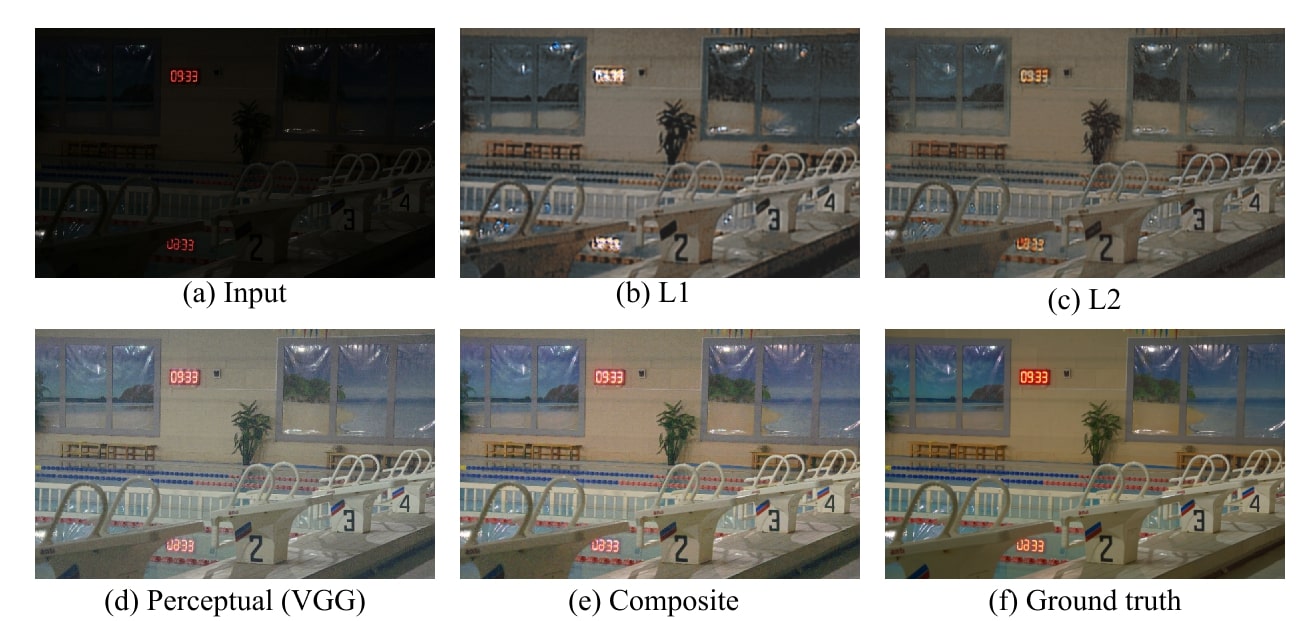}
    \caption{Visual comparison of the proposed model's performance on the LOL dataset using four loss functions, reported without post-processing. The visual result achieved by our model trained with the (e) composite loss function exhibits the highest quality.}
    \label{fig:fig12}
\end{figure}

In pursuit of a comprehensive evaluation, we embarked on a series of experiments to scrutinize the efficacy of various widely used loss functions in the context of our proposed model. The quantitative evaluation of the proposed model trained with different loss functions is presented in Table \ref{tab:ablation_loss}, and visual comparisons are provided in Figure \ref{fig:fig12} to provide a more tangible understanding of the effects of loss functions on the perceptual quality of the generated images.

The L1 loss function resulted in a weak PSNR of 11.335 and SSIM of 0.506. Figure \ref{fig:fig12}(b) visually indicates its inferior perceptual quality as well. A noticeable performance improvement becomes apparent with the adoption of the L2 loss function, yielding a notable PSNR of 19.325 and a significantly improved SSIM of 0.830. Transitioning to the Perceptual loss function (utilizing VGG-based perceptual similarity), we observe that the PSNR remains high at 19.209, albeit with a marginal drop, while the SSIM stabilizes at 0.810. Despite the numerical decline in evaluations, Figure \ref{fig:fig12}(d) highlights enhanced image quality, resulting in a more natural look and fewer artifacts. Our proposed composite loss, combining the L2 and VGG losses, outperforms individual loss functions. It achieves an average PSNR of 19.569 and an average SSIM of 0.820, demonstrating superior performance compared to other loss functions.

In conclusion, the ablation study on loss functions demonstrates the importance of choosing an appropriate loss function for low-light image enhancement tasks. While traditional pixel-level losses, such as L1 and L2, contribute to fidelity, incorporating perceptual losses, such as the VGG loss, can significantly enhance the visual quality. Our proposed composite loss, combining both L2 and VGG losses, achieves the best trade-off between fidelity and perceptual quality, resulting in superior performance in terms of both PSNR and SSIM metrics.
\begin{table}
    \caption{Comparing the proposed model's performance on the LOL dataset using four loss functions, reported without post-processing.}
    \centering
    \begin{tabular}{ccc}
    \toprule
    \textbf{Loss function} & \textbf{Avg. PSNR} $\uparrow$ & \textbf{Avg. SSIM} $\uparrow$\\
    \midrule
    L1 & 11.335 & 0.506\\
    L2 & 19.325 & 0.830\\
    Perceptual (VGG) & 19.209 & 0.810\\
    Composite (L2 + VGG)& \textbf{19.569} & \textbf{0.820}\\
    \bottomrule
    \end{tabular}
    \label{tab:ablation_loss}
\end{table}

\subsubsection{Study on the effects of post-processing}
We conducted an ablation study to examine the impact of post-processing on the outcomes of our model, and the results are summarized in Table \ref{tab:ablation_pp}. The comparison of average PSNR, SSIM, and LPIPS scores before and after the post-processing phase provides valuable insights into the effectiveness of this additional step.

Regarding numerical metrics, marginal decreases are observed in SSIM and LPIPS scores following the post-processing step. SSIM score, with a value of 0.820 before post-processing, decreases to 0.816 after post-processing. Similarly, the LPIPS score goes from 0.164 to 0.167, representing a slight increase in perceptual dissimilarity. In contrast, the PSNR metric improves from 19.569 to 20.102 after post-processing, indicating a reduction in pixel-level noise and an increase in visual similarity to the ground truth. 

\begin{figure}
    \centering
    \includegraphics[width=\linewidth]{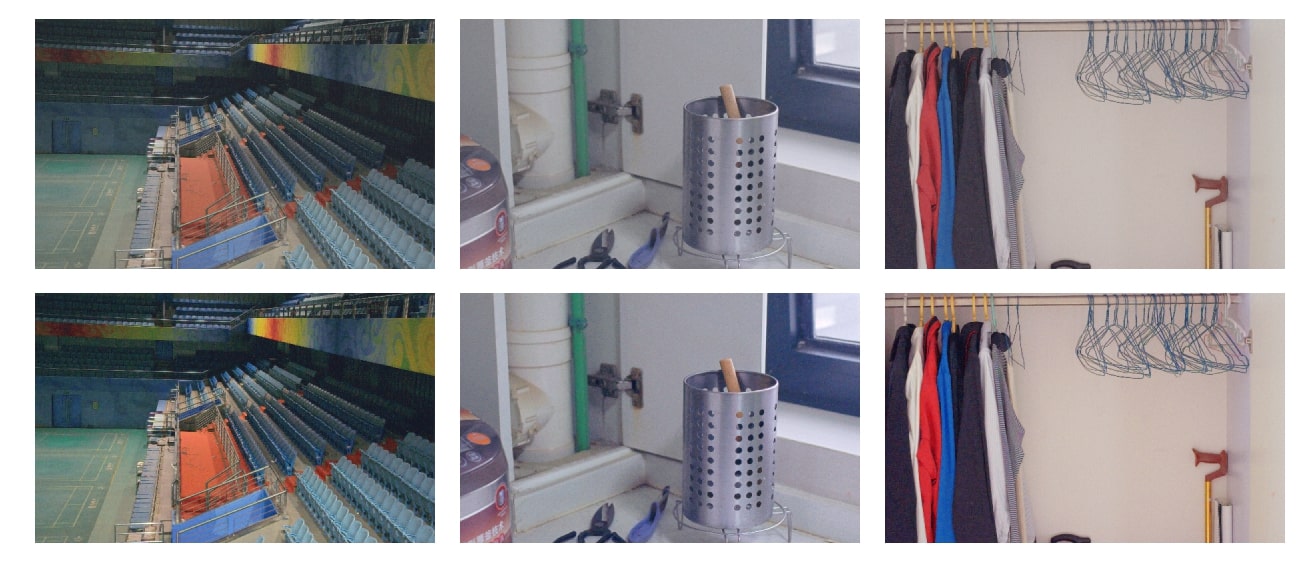}
    \caption{Visual comparison of the proposed model's outputs before (top row) and after (bottom row) the post-processing.}
    \label{fig:fig13}
\end{figure}

However, it is essential to emphasize that the significance of the post-processing phase goes beyond these numerical measures. The visual impact of the post-processing step is vividly evident in Figure \ref{fig:fig13}. The perceptual enhancements resulting from post-processing are readily discernible, with the generated images exhibiting more vibrant colors, better-defined edges, and an overall visual appeal. Furthermore, it is noticeable that our study employed a relatively traditional and simple post-processing approach, leaving room for more sophisticated techniques to be explored in future research.

\begin{table}
    \caption{Comparison of PSNR, SSIM, and LPIPS scores before and after post-processing on the LOL test dataset.}
    \centering
    \begin{tabular}{ccc}
        \toprule
        \multirow{2}{*}{\textbf{Evaluation metrics}} & \multicolumn{2}{c}{\textbf{Post-processing}} \\
        \cmidrule{2-3}
        & \textbf{Before} & \textbf{After} \\
        \midrule
        PSNR $\uparrow$ & 19.569 & \textbf{20.102}\\
        SSIM $\uparrow$ & \textbf{0.820} & 0.816\\
        LPIPS $\downarrow$ & \textbf{0.164} & 0.167\\
        \bottomrule
    \end{tabular}
    \label{tab:ablation_pp}
\end{table}

\section{Conclusion}
In this paper, we introduce the Convolutional Dense Attention-guided Network (CDAN) as a novel and powerful solution for enhancing low-light images. Our approach combines the power of autoencoders with key architectural components, including convolutional blocks, dense blocks, attention modules, and skip connections. To address the degradation challenge posed by the encoder, we employed a multi-branch strategy for feature transfer from the encoder to the decoder. This strategy effectively alleviates degradation issues and enables efficient information flow throughout the network. We employ a composite loss function during training, which combines L2 and VGG losses. This innovative strategy produces results that are both visually pleasing and supported by state-of-the-art numeric measurements. Additionally, our post-processing phase improves color balance and contrast, leading to performance that aligns with state-of-the-art models on benchmark datasets. Our proposed model proves adept at addressing variations in lighting conditions, yielding well-lit images with significant potential for diverse computer vision tasks, such as object detection and recognition within challenging low-light environments. However, it is worth highlighting that there are still challenges in areas that include intensely illuminated regions, where our model tends to over-enhance those parts. This direction warrants further investigation and refinement in future work to ensure robustness across an even broader spectrum of lighting conditions.
\bibliographystyle{unsrt}  
\bibliography{references}  

\end{document}